\documentclass[letterpaper, 10 pt, conference]{ieeeconf}  %

\usepackage{balance}
\usepackage{booktabs}
\usepackage{threeparttable}
\usepackage{siunitx}
\usepackage{cite}

\IEEEoverridecommandlockouts                              %

\usepackage{url}
\usepackage{graphicx} %
\usepackage[hybrid]{markdown} %

\makeatletter
\let\NAT@parse\undefined
\makeatother

\usepackage{cite}
\usepackage[colorlinks,allcolors=magenta]{hyperref} %
\usepackage[all]{hypcap}

\title{\LARGE \bf
RootSLAM: Square Root Bundle Adjustment for Visual-Inertial SLAM
}

\author{Tom\'as Santucci$^{*1}$, Mateo de Mayo$^{*2}$, Daniel Cremers$^2$, Taih\'u Pire$^{1,3}$%
\thanks{*Equal contribution.}%
\thanks{$^{1}$Tom\'as Santucci and Taih{\'u} Pire are with the National University of Rosario, Rosario, Argentina.
        {\tt\small tsantucci@dcc.fceia.unr.edu.ar}}%
\thanks{$^{2}$Mateo de Mayo and Daniel Cremers are with the Technical University of Munich and the Munich Center for Machine Learning, Munich, Germany.
        {\tt\small \{mateo.demayo, cremers\}@tum.de}}%
\thanks{$^{3}$Taih\'u Pire is also with the CIFASIS (CONICET-UNR), Rosario, Argentina.
    {\tt\small pire@cifasis-conicet.gov.ar}}%
}

\newread\imgstream
\immediate\openin\imgstream=imagedata.in
\makeatletter
\def\new@kvginclip#1{}
\def\new@kvgintrim#1{}
\let\old@kvginclip\KV@Gin@clip
\let\old@kvgintrim\KV@Gin@trim
\let\oldincludegraphics\includegraphics
\providecommand{\includegraphics}{}
\renewcommand{\includegraphics}[2][]{%
  \immediate\read\imgstream to \src
  \immediate\read\imgstream to \removecrop
  \ifnum\removecrop=1
      \let\KV@Gin@clip\new@kvginclip
      \let\KV@Gin@trim\new@kvgintrim
  \fi
  \oldincludegraphics[#1]{\src}%
  \let\KV@Gin@clip\old@kvginclip
  \let\KV@Gin@trim\old@kvgintrim}
\makeatother

\begin{document}

\maketitle
\thispagestyle{empty}
\pagestyle{empty}

\begin{abstract}

Efficient perception is a prerequisite for the small, low-power devices targeted by robotics, mixed reality, autonomous driving, and drone applications, all of which rely on Simultaneous Localization and Mapping (SLAM) to estimate their motion within their surroundings. Learning-based SLAM systems achieve state-of-the-art accuracy but are demanding in compute, whereas classical geometry-based methods retain a strong efficiency advantage with little to no loss in accuracy. In this work we present RootSLAM, a fully-featured visual-inertial SLAM system for multi-camera setups built on Basalt. RootSLAM achieves accuracy competitive with the state of the art while running up to 20 times faster, facilitating the deployment of perception systems in resource constrained environments. We adapt the square-root bundle adjustment formulation from structure from motion into the multi-camera visual-inertial SLAM setting through a novel reformulation that leverages the system's pose-graph, letting the global optimizer run in single-precision floating point, thus doubling its performance. We provide a thorough benchmark against the state of the art together with an ablation study of our contributions. Our new implementation, including its map management and loop detection module, pose-graph optimizer, and global mapping components, will be open-sourced.

\end{abstract}

\section{INTRODUCTION}

In recent years, the need for computing devices to better understand their environments has become increasingly important. Many disciplines from robotics, mixed reality, drone navigation, among others, all share a common goal for devices to understand and \textit{perceive} their surroundings. This has been driving ideas like the spatial computing paradigm \cite{delmericoSpatialComputingIntuitive2022} or the perception engineering discipline \cite{lavalleVirtualRealityEmerging2024a}, to categorize a similar family of problems and solutions that aim to achieve perception capabilities on a computing platform.

While the applications and tasks these systems need to perform are widely different, all of them share a common need of understanding their environment. A foundational prerequisite for this is the problem of localizing themselves and their movements within their surroundings. It is in this context that Simultaneous Localization and Mapping (SLAM) has come to play an important role. In particular, visual-inertial odometry (VIO) and SLAM (VI-SLAM) have been shown to offer a particularly good combination of sensors due to their complementarity and low cost, especially when coupled with two or more cameras to reduce depth ambiguity and increase the tracking field of view \cite{eckenhoffMIMCVINSVersatileResilient2021}. Learning-based methods for SLAM have shown promise in terms of state-of-the-art accuracy but they demand significantly higher compute compared to classical methods due to the implicit overhead in approximating geometry through neural networks and the use of GPUs. Another downside shown in the recent \cite{cioffiWhyDoesDeep2026}, is that most of the accuracy gains in current learning-based systems come from the improved feature tracks and the proper handling of their uncertainty. The same work shows that when replicating these in a classical geometry-based pipeline like ORB-SLAM3 \cite{camposORBSLAM3AccurateOpenSource2021a}, the classical systems outperform the learned ones. Therefore, classical geometry-based systems have a non-trivial edge in performance with little to no losses in accuracy.

\begin{figure}[t]
   \centering
   \includegraphics[width=\columnwidth]{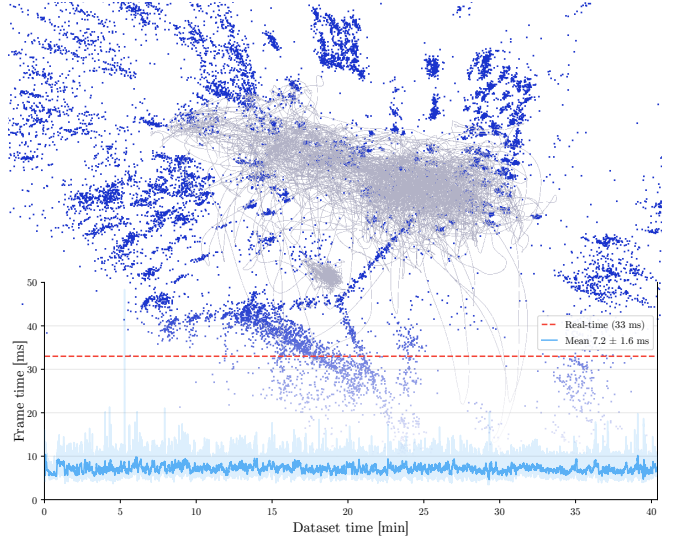}
   \caption{RootSLAM running on the long sequence MOO12. Top: the resulting map, with the estimated trajectory in gray and the landmarks in blue. Bottom: moving average of the per-frame processing time throughout the sequence, which stays under the real-time budget of 33\,ms for 99.99\% of the frames, averaging $7.2$\,ms with a standard deviation of $1.6$\,ms. Run on an Intel Core Ultra 9 285H.}
   \label{fig:teaser}
\end{figure}

Runtime and energy efficiency are heavily correlated properties, thus providing faster systems allows for smaller, cheaper, and cooler devices with perception capabilities. In this work we introduce \textit{RootSLAM}, a fully-featured VI-SLAM system based on Basalt \cite{usenkoBasaltVisualInertialMapping2020} with the modifications from \cite{demayoMonadoSLAMDataset2025}, that achieves competitive accuracy numbers while being up to $20\times$ faster than state-of-the-art solutions like ORB-SLAM3 \cite{camposORBSLAM3AccurateOpenSource2021a} and OKVIS2 \cite{leuteneggerOKVIS2RealtimeScalable2022}. Figure~\ref{fig:teaser} shows the system running on a 42-minute sequence, where it keeps its per-frame processing time well under the real-time budget throughout the session. In RootSLAM we leverage the square root formulation from Demmel et al. \cite{demmelSquareRootBundle2021} which allows us to optimize large bundle adjustment problems with single precision arithmetic. We adapt it from structure from motion into the multi-camera SLAM setting with a novel reformulation that accounts for our pose-graph. This enables our global optimizer to work in single precision accuracy, effectively doubling its performance.

In summary, the contributions of this work are as follows:

\begin{itemize}
    \item A new highly-efficient, low-latency, and multi-camera VI-SLAM system that is up to $20\times$ faster than the current state-of-the-art with competitive accuracy.
    \item A novel square root formulation of \cite{demmelSquareRootBundle2021} for the global optimization step of multi-camera VI-SLAM, allowing us to perform large map optimizations in 32-bit floating point precision, instead of the usual 64-bit, doubling our performance.
    \item A thorough benchmark against the state of the art and ablation study show-casing the improvements of our different components.
\end{itemize}

Our complete implementation will be open-sourced, including its long-term map management, loop detection and global mapping components.

\section{RELATED WORK}

The earliest approaches to visual-inertial odometry utilized Kalman filtering theory for state estimation. MSCKF \cite{mourikisMultiStateConstraintKalman2007} introduced an algorithm based on an extended Kalman filter (EKF) for VIO systems, removing the 3D point positions from the state vector. This enabled linear computational complexity in the number of observed features. ROVIO \cite{bloeschRobustVisualInertial2015} took an alternative approach using an iterated extended Kalman filter (IEKF), employing photometric error over image patches. Recent filter-based methods such as OpenVINS \cite{genevaOpenVINSResearchPlatform2020a} and sqrtVINS \cite{pengSqrtmathbfVINSRobust2025} have achieved remarkable results but still lag behind optimization-based systems in terms of accuracy metrics like absolute trajectory error (ATE).

Recent literature has explored the use of neural networks in the SLAM pipeline, such as DROID-SLAM \cite{teedDROIDSLAMDeepVisual2021} and Deep Patch Visual SLAM \cite{lipsonDeepPatchVisual2025}, the latter being significantly faster than its predecessors. These approaches have achieved state-of-the-art accuracy but, as shown in \cite{cioffiWhyDoesDeep2026}, their advantages seem to come from their front-end modules, which also perform well when coupled with classical geometry-based methods. In contrast, the system proposed in this work is computationally more efficient, requires no training stage, runs fully on CPU, and is better suited to resource-constrained platforms.

Unlike filter-based methods, sliding-window least-squares optimization has demonstrated higher accuracy and computational efficiency. This approach has become the dominant paradigm in VI-SLAM systems, such as PTAM \cite{kleinParallelTrackingMapping2007} and ORB-SLAM \cite{mur-artalORBSLAMVersatileAccurate2015}. OKVIS \cite{leuteneggerKeyframebasedVisualInertial2015} formulated visual-inertial fusion as a sliding-window nonlinear least-squares problem, incorporating reprojection errors and IMU preintegration. Basalt \cite{usenkoBasaltVisualInertialMapping2020} is a keyframe-based optimization VIO method, utilizing FAST \cite{rostenMachineLearningHighSpeed2006} for feature detection and sparse KLT optical flow~\cite{Lucas-1981-15101} for feature tracking. The system optimizes a window of keyframes with a marginalization prior, maintaining a smooth estimate. Furthermore, the data marginalized by VIO can be processed to construct a global map and recover new point correspondences, which are refined through global bundle adjustment over all poses and landmarks. Our work replaces this offline step, providing a much more accurate mapping alternative that is better suited for real-time deployment.

Kimera \cite{rosinolKimeraOpenSourceLibrary2020a} is a metric-semantic VI-SLAM system that incorporates loop closure via DBoW2 \cite{galvezlopez2012dbow2} and pose-graph optimization, and also integrates 3D mesh reconstruction and semantic labeling. Snake-SLAM \cite{ruckertSnakeSLAMEfficientGlobal2021} decouples the optimization, solving the IMU state and visual bundle adjustment as separate subproblems, reducing computational complexity without compromising accuracy. ORB-SLAM3 \cite{camposORBSLAM3AccurateOpenSource2021a} unifies multiple input modalities with support for the IMU, short-, mid-, and long-term data association, and an Atlas multi-map system for managing disconnected maps. It has become one of the most accurate and widely adopted VI-SLAM systems in the literature. OKVIS2 \cite{leuteneggerOKVIS2RealtimeScalable2022} extends OKVIS with a persistent map, efficient loop closure, and marginalization, remaining among the most accurate state-of-the-art systems, together with ORB-SLAM3, on standard benchmarks such as EuRoC \cite{burriEuRoCMicroAerial2016a} and TUM-VI \cite{schubertBasaltTUMVI2018}.

Global bundle adjustment for map refinement is an expensive stage in an optimization-based SLAM, with the cost dominated by solving the normal equations of a nonlinear least-squares problem \cite{triggsBundleAdjustmentModern2000a}. Classical approaches form the normal equations, which squares the condition number of the problem, necessitating double-precision arithmetic for numerical stability.

Square-root formulations avoid this issue by operating on the Jacobian through orthogonal (QR) factorization. This concept originates in the square-root information filter of state estimation and was introduced to SLAM by Square Root SAM \cite{dellaertSquareRootSAM2006}, which solves the full smoothing problem via QR or Cholesky factorization of the measurement Jacobian. iSAM \cite{kaessISAMIncrementalSmoothing2008} and iSAM2 \cite{kaessISAM2IncrementalSmoothing2012} adapted this idea for incremental schemes, updating the square-root factor with Givens rotations and organizing variables in the Bayes tree for efficient partial relinearization.

Demmel et al. \cite{demmelSquareRootBundle2021} recently revisited this concept for visual bundle adjustment, marginalizing each landmark by nullspace projection through local QR decomposition. This method improves numeric stability, allowing large-scale structure-from-motion problems to be solved in single-precision floating point with accuracy comparable to double-precision Schur solvers and with significant runtime reduction. The follow-up \cite{demmelSquareRootMarginalization2021} extends this concept to the sliding-window setting, deriving a square-root form of the marginalization prior used by windowed visual-inertial estimators such as Basalt. Our work builds on this foundation, adapting its nullspace landmark marginalization of \cite{demmelSquareRootBundle2021} to the global optimization stage of a multi-camera visual-inertial SLAM system, reformulating it around the pose-graph to enable efficient single-precision computation by keeping a square-root of the reduced camera system's hessian.

Summarizing the literature, OKVIS2 and ORB-SLAM3 achieve the highest state-of-the-art accuracy \cite{krishnanBenchmarkingEgocentricVisualInertial2025a}, but as demonstrated in \cite{demayoMonadoSLAMDataset2025}, OKVIS2 is significantly heavy in terms of compute times while ORB-SLAM3 significantly degrades relative trajectory error (RTE), which translates into noisy and non-smooth trajectory estimates, particularly important for XR and high-speed tracking. Basalt, in contrast, is a robust VIO system that attains state-of-the-art RTE while running more efficiently, even without a persistent map. However, it lacks a loop closure module, preventing it from achieving good absolute trajectory error (ATE). Our work builds on Basalt and extends it with a persistent map, loop closure, and global optimization modules, maintaining its efficiency while surpassing the ones from full VI-SLAM systems like OKVIS2 and ORB-SLAM3, and significantly closing the gap in absolute trajectory accuracy.

\section{METHOD}

\begin{figure*}[!htbp]
    \centering
    \includegraphics[width=1\linewidth]{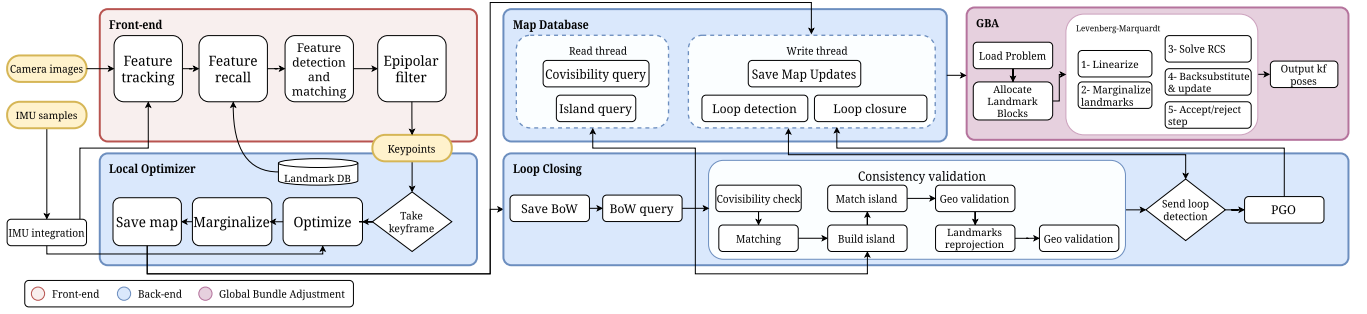}
    \caption{Architecture of RootSLAM. We introduce the map database, loop closing and global bundle adjustment components and add them on top of Basalt's front-end and local optimizer.}
    \label{fig:system-pipeline}
\end{figure*}

RootSLAM is built on top of Basalt \cite{usenkoBasaltVisualInertialMapping2020} with the modifications introduced in \cite{demayoMonadoSLAMDataset2025}. This is an efficient multi-camera VIO system, but it only supports short- and mid-term data association, and its estimate therefore drifts over long sessions. We extend it into a fully-featured VI-SLAM system with the three components shown in Figure~\ref{fig:system-pipeline}: a persistent map that outlives the sliding window of the local optimizer, a loop closing module that provides long-term data association, and a global optimization stage that replaces the non-linear factor recovery of Basalt. All three exploit map information to improve accuracy while preserving the efficiency that characterizes the original system.

We write $T_{ab} \in SE(3)$ for the transform that maps points from frame $b$ into frame $a$, and $w$ for the world frame, so the pose of a keyframe $a$ is $T_{wa}$. We abbreviate it as $T_a$ in the pose-graph optimizer. The global bundle adjustment inherits the world-to-camera convention of RootBA, and there $T_a$ abbreviates $T_{aw}$ instead.

\subsection{Persistent Map}

We implement a persistent map, which holds the keyframes and landmarks that the local optimizer marginalizes out of its sliding window. Alongside them we maintain a set of graphs, including a covisibility graph, kept up to date at all times and inspired by the \textit{Essential Graph} of ORB-SLAM~\cite{mur-artalORBSLAMVersatileAccurate2015}. With them we form a lightweight representation of the map, that we later optimize during pose-graph optimization.

To bound the map, we need to cull keyframes. On marginalization, we discard those whose observations are over 90\% redundant with three or more other keyframes. If its size grows over 1500 keyframes, we drop the 30 worst-ranked by a quality and covisibility-redundancy score. Culled keyframes take their hosted landmarks with them, and all these thresholds can be configured.

\subsection{Loop Detection and Closing}

We provide long-term data association with an efficient loop closing module that receives, at keyframe rate, a pointer to its images and their keypoints. It runs asynchronously on a separate thread, so neither the local optimizer nor the map ever block waiting for it.

\begin{figure}[t]
   \centering
   \includegraphics[width=\columnwidth]{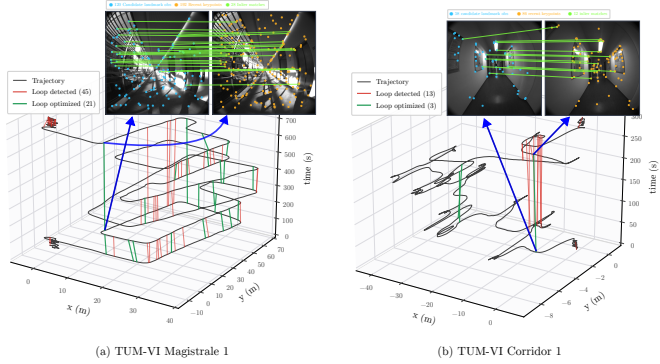}
   \caption{Loop closures found by RootSLAM on (a)~TUM-VI magistrale~1 and (b)~corridor~1. The trajectory is drawn with time on the vertical axis, so every near-vertical segment links two visits to the same place: in red the loops that were detected and validated, and in green the subset whose estimated drift exceeded 10\,cm and therefore triggered a pose-graph optimization. The insets show the 3D-2D correspondences of one such loop, with the observations of the candidate landmarks, the keypoints of the new keyframe, and the inlier matches that survive the geometric validation.}
   \label{fig:loopclosure}
\end{figure}

For candidate detection we use HashBoW \cite{hashbow2024stricker}, due to its high speed. We compute ORB~\cite{rublee2011orb} descriptors over the corners already detected by the front-end, build the corresponding bag-of-words vector, and query the database with it. After this, every candidate goes through a chain of consistency validations. Among them, we do brute-force matching and landmark reprojection, that establish 3D-2D correspondences, and a final geometric validation, run with gP3P \cite{kneip2013gp3p} from the OpenGV library \cite{kneip2014opengv}.

A validated loop is reported to the persistent map, which reacts according to the magnitude of the estimated drift. Below 10 cm it simply fuses the landmarks validated as inliers. Above it, the map hands every keyframe pose to the loop closing thread, together with the pose-graph edges. The thread optimizes it and returns the corrected poses to the map, which synchronizes them before fusing their landmarks. Figure~\ref{fig:loopclosure} shows the loops that the module detects and validates on two TUM-VI sequences, and the subset of them that exceeds the drift threshold and therefore triggers a pose-graph optimization.

Writing $T_a = T_{wa}$ for the pose of keyframe $a$ and $\bar{T}_{ab}$ for the measured relative pose of the edge $(a,b)$, we minimize
\begin{equation}
\min_{\{T_a\}} \sum_{(a,b) \in \mathcal{E}} \left\lVert
  \log\!\big( \hat{T}_{ab}^{-1}\, \bar{T}_{ab} \big)
\right\rVert^2 ,
\label{eq:pgo}
\end{equation}
where $\hat{T}_{ab} = T_a^{-1} T_b = T_{aw}\,T_{wb}$ is the relative pose predicted by the current estimates. The measurements are taken from the current estimates for the existing edges and from the gP3P pose for the ones the loop introduces. We solve it with Levenberg-Marquardt in Ceres \cite{ceressolver}, over a quaternion manifold and with a sparse Cholesky solver. The keyframes still active in the local optimizer's sliding window are held fixed, which anchors the gauge and confines the correction to the rest of the map.

\subsection{Global Optimization}

To refine the estimation, we close the pipeline with a global bundle adjustment step over the whole map, jointly optimizing every keyframe pose and landmark position.

We build it on RootBA \cite{demmelSquareRootBundle2021}. We first summarize its formulation, and then describe the modifications that carry it from structure from motion (SfM) into a multi-camera visual-inertial setting. We remove the intrinsic parameters from the state, parameterize it with rig poses, add support for more camera models, and introduce a pose-graph regularizer.

\textbf{Square-root bundle adjustment.} RootBA splits the state $x$ into a pose part $x_p$ and a landmark part $x_l$, the latter holding the 3D coordinates of $n_l$ points. The pose part is further subdivided into the camera poses $x_c$ and the intrinsic parameters $x_i$ (focal length and radial distortion coefficients), so each of the $n_p$ cameras contributes $d_p = 9$ parameters and the full state has length $d_p n_p + 3 n_l$. The problem consists of minimizing the energy
\begin{equation}
E(x_c, x_i, x_l) = \sum_{(i,j) \in \mathcal{O}} \lVert r_{ij}(x_c, x_i, x_l) \rVert ^2,
\label{eq:ba_energy}
\end{equation}
where $(i,j) \in \mathcal{O}$ indicates that landmark $j$ is observed in frame $i$. The Jacobians and residuals of each landmark are condensed into a dense matrix called a \textit{landmark block}, and the energy is minimized by the Levenberg-Marquardt algorithm, which linearizes $r(x)$ around the current estimate $x^0 = (x_p^0, x_l^0)$ and solves a damped linear problem at every iteration. Writing the undamped case for brevity, that problem is equivalent to Gauss-Newton and it reads
\begin{equation}
\min_{\Delta x_p, \Delta x_l} \left\lVert r + \left( J_p \; J_l \right) \left( \begin{array}{c} \Delta x_p \\ \Delta x_l \end{array} \right) \right\rVert^2.
\label{eq:ba_linearized}
\end{equation}

We can QR-decompose the landmark Jacobian
\begin{equation}
J_l= (Q_1 \; Q_2) \left( \begin{array}{c} R_1 \\ 0 \end{array} \right),
\label{eq:ba_qr}
\end{equation}
with $R_1$ upper triangular and $Q_2$ an orthonormal basis of the left nullspace of $J_l$. Projecting Equation~\eqref{eq:ba_linearized} onto that nullspace eliminates the landmark variables and yields the equivalent reduced camera system
\begin{equation}
\min_{\Delta x_p} \left\lVert Q_2^T r + Q_2^T J_p \Delta x_p \right\rVert^2.
\label{eq:ba_linearized_marg}
\end{equation}
Once it has been solved for the optimal pose update $\Delta x_p^*$, the corresponding landmark update follows by back-substitution,
\begin{equation}
\Delta x_l^* = -R_1^{-1} \left( Q_1^T r + Q_1^T J_p \Delta x_p^* \right).
\label{eq:ba_landmark_backsub}
\end{equation}
Note that the normal equations are never formed. The landmarks are marginalized by an orthogonal transformation, which is what preserves the conditioning of the problem in single-precision arithmetic.

\textbf{Removal of intrinsic parameters.} In SfM the intrinsics are unknown, as the images may come from different and uncalibrated cameras. In SLAM, in contrast, the calibration is usually known and we assume it fixed. Optimizing $x_i$ may be harmful then, because we could absorb pose errors into the calibration instead of correcting $x_c$. We therefore remove $x_i$ from the state, reducing the energy to $E(x_c, x_l)$ and the per-camera dimension from $d_p = 9$ to $d_p = 6$. In the landmark blocks this amounts to dropping the columns associated with $x_i$, which leaves the structure of the QR factorization untouched and spares the computation of the Jacobians with respect to the intrinsic parameters. We further generalized the implementation, which was restricted to a pinhole model with radial distortion, to support the arbitrary camera models available in Basalt.

\begin{figure}[!htbp]
    \centering
    \includegraphics[width=\linewidth]{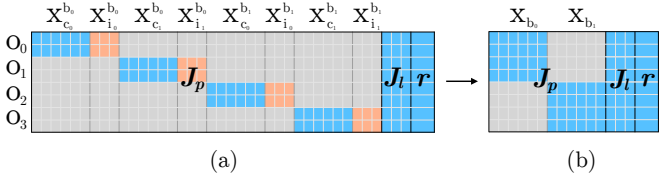}
    \caption{Structure of a landmark block. (a)~Original RootBA: per-camera state of dimension $d_p = 9$, including the intrinsic parameters $x_i$ in orange. Rows correspond to the observations of the landmark. (b)~RootSLAM removes the intrinsics, so $d_p=6$, and accumulates observations from multiple cameras of the rig in the same body pose variable $x_b$.}
    \label{fig:landmark_block}
\end{figure}

\textbf{Rig pose optimization.} In a keyframe-based multi-camera system, each keyframe is a rig of cameras $c_k$ with $k \in \{0, 1, \ldots \}$, whose transformations from the body frame $T_{c_k b} \in SE(3)$ are given by the extrinsic calibration. We assume that these transformations remain constant across keyframes. Here $T_{c_k} = T_{c_k w}$ and $T_b = T_{bw}$, the world-to-body transform of the keyframe, whose body frame we set to match the IMU, so that $T_{c_k} = T_{c_k b}\cdot T_b$. Storing one independent transform per camera per keyframe in $x_c$, as the original formulation does, would let the optimizer violate this rigidity, so we re-parameterize the state with the body poses $x_b$ instead. This requires accounting for new Jacobians. Since $\Delta x_{c_k} = \mathrm{Ad}(T_{c_k b}) \Delta x_b$, the chain rule gives
\begin{equation}
\frac{\partial r}{\partial \Delta x_b} = \frac{\partial r}{\partial \Delta x_{c_k}} \cdot \frac{\partial \Delta x_{c_k}}{\partial \Delta x_b} = \frac{\partial r}{\partial \Delta x_{c_k}}\cdot \mathrm{Ad}(T_{c_k b}).
\label{eq:rig_chain_rule}
\end{equation}
As a consequence, a single landmark block accumulates the observations of every camera of a rig in the one column that corresponds to its body pose. Figure~\ref{fig:landmark_block} illustrates the joint effect of these two modifications on the structure of a landmark block. From here on the pose part of the state is the set of body poses, $x_p = x_b$.

\textbf{Pose-graph regularizer.} Bundle adjustment only constrains keyframes that observe common landmarks, whereas the pose-graph maintained by the map also relates them through the odometry and through previously closed loops. Our reformulation exploits it as an additional source of constraints, adding one relative pose residual per edge in a form that keeps the reduced camera system in square-root form. With $T_a = T_{aw}$ as above, we denote by $\bar{T}_{ab}$ the measured relative pose of the edge $(a,b)$ and by $\hat{T}_{ab} = T_a T_b^{-1} = T_{aw}\,T_{wb}$ its current estimate. The residual of the edge is
\begin{equation}
r_{ab} = S_{ab}\, e_{ab}, \qquad e_{ab} = \log\!\big( \bar{T}_{ab}\, \hat{T}_{ab}^{-1} \big) \in \mathbb{R}^6,
\label{eq:pg_residual}
\end{equation}
where $S_{ab}$ is the square root of the information matrix of the edge. Equation~\eqref{eq:pgo} expresses the same error in the frame of $\hat{T}_{ab}$ instead, to match the right perturbation of its solver. While reprojection residuals from bundle adjustment are measured in pixels, pose-graph residuals are measured in meters and radians. We use $S_{ab}$ to bring both to a common scale and make their sum meaningful. Adding these terms, the energy from Equation~\eqref{eq:ba_energy} becomes
\begin{equation}
E(x_p, x_l) = \sum_{(i,j) \in \mathcal{O}} \lVert r_{ij} \rVert ^2 + \sum_{(a,b) \in \mathcal{E}} \lVert r_{ab} \rVert ^2,
\label{eq:ba_energy_pg}
\end{equation}
with $\mathcal{E}$ the set of edges of the pose-graph. Under a left perturbation model, the only non-zero blocks of the Jacobian of $r_{ab}$ are those of the two keyframes it relates,
\begin{equation}
\frac{\partial r_{ab}}{\partial \Delta x_a} = -S_{ab}\, \mathcal{J}(e_{ab}), \quad
\frac{\partial r_{ab}}{\partial \Delta x_b} = S_{ab}\, \mathcal{J}(e_{ab})\, \mathrm{Ad}\!\big(\hat{T}_{ab}\big),
\label{eq:pg_jacobians}
\end{equation}
where $\mathcal{J}$ denotes the inverse right Jacobian of $SE(3)$, evaluated at $e_{ab}$.

These residuals involve no landmark, so there is nothing to marginalize and their rows need no nullspace projection. We append them directly below the $Q_2^T J_p$ rows of the reduced camera system in Equation~\eqref{eq:ba_linearized_marg}, which, with the damping term now made explicit, becomes
\begin{equation}
\min_{\Delta x_p} \left\lVert
  \left( \begin{array}{c} Q_2^T J_p \\ J_{\mathrm{pg}} \\ \sqrt{\lambda}\,I \end{array} \right)
  \Delta x_p
  +
  \left( \begin{array}{c} Q_2^T r \\ r_{\mathrm{pg}} \\ 0 \end{array} \right)
\right\rVert^2,
\label{eq:rcs_pg}
\end{equation}
where $J_{\mathrm{pg}}$ and $r_{\mathrm{pg}}$ stack Equations~\eqref{eq:pg_residual} and \eqref{eq:pg_jacobians} over all the edges of $\mathcal{E}$, each row block being zero except for the two $6 \times 6$ columns of the keyframes it relates. The pose-graph constraints thus enter as extra rows of the square-root factor rather than as blocks of a Hessian, so the reduced camera system is never squared and the global optimization runs entirely in single-precision floating point.

\section{RESULTS}

We evaluate on the 103 sequences of the Monado SLAM Dataset (MSD) \cite{demayoMonadoSLAMDataset2025}, EuRoC \cite{burriEuRoCMicroAerial2016a} and TUM-VI \cite{schubertBasaltTUMVI2018}. For space, we group sequences by category and report averaged numbers, with the number per group in the \textbf{\#} column.

We take Basalt as our baseline and compare against OKVIS2 \cite{leuteneggerOKVIS2RealtimeScalable2022} and ORB-SLAM3 \cite{camposORBSLAM3AccurateOpenSource2021a} as state-of-the-art references, both consume multi-camera frames and IMU samples, and run fully smoothed and non-causal for maximum accuracy. ORB-SLAM3 is the only system unable to use the 4 cameras of the 15 MGO sequences from MSD, so we run it on the two front cameras only. The rest of the sequences offer only two cameras.

We report root mean squared absolute and relative trajectory errors (RMSE ATE and RTE) in centimeters, and success, the percentage of camera frames for which a system produced an estimate. We use $SE(3)$ alignment for the ATE, and $\Delta = 6$ frames for the RTE, like in \cite{demayoMonadoSLAMDataset2025}. All the numbers are non-causal, computed over the final trajectory reported once the whole sequence has been processed.

Table~\ref{tab:ablation-accuracy} adds our contributions one by one on top of Basalt's VIO-only baseline (\textbf{BSL}). The persistent map with loop closing (\textbf{+LC}), the square-root global bundle adjustment on top of it (\textbf{+GBA}), and its pose-graph terms, which complete RootSLAM (\textbf{RS}). Long-term data association dominates the gain in global accuracy. Loop closing alone already divides the ATE of the baseline by up to $6\times$, from 23.6 to 4.0\,cm on MIP, and the complete system refines it down to a $10\times$ reduction on that same group. RootSLAM improves the ATE of Basalt in all the twelve groups.

Global bundle adjustment on its own improves ATE but degrades RTE with respect to \textbf{+LC} (1.48 to 1.88\,cm on MIO, 1.33 to 2.11\,cm on EV1), as it redistributes the error to fit the map globally at the expense of local consistency. Adding the pose-graph residuals of Equation~\eqref{eq:pg_residual} recovers it in every group while preserving the ATE improvement. The complete system also improves the RTE of the VIO baseline in eight of the twelve groups and degrades it by at most 0.25\,cm in the rest (excluding the outlying TO), so global accuracy is gained without sacrificing the smoothness of the trajectory, the property that matters most for XR and high-speed tracking.

\begin{table}[ht]
\centering
\resizebox{\columnwidth}{!}{%
\begin{threeparttable}
\caption{Ablation study over the components of RootSLAM}
\label{tab:ablation-accuracy}
\setlength{\tabcolsep}{2.5pt}
\begin{tabular}{@{}l r rrrr | rrrr@{}}
\toprule
& & \multicolumn{4}{c|}{\textbf{ATE [cm]}} & \multicolumn{4}{c}{\textbf{RTE [cm]}} \\
\cmidrule(lr){3-6} \cmidrule(lr){7-10}
\textbf{Seq.}
& \textbf{\#}
& \textbf{BSL} & \textbf{+LC} & \textbf{+GBA} & \textbf{RS}
& \textbf{BSL} & \textbf{+LC} & \textbf{+GBA} & \textbf{RS} \\
\midrule
MIO & 16 & 33.0 & 9.3 & 8.3 & \textbf{7.0} & 1.38 & 1.48 & 1.88 & \textbf{1.26} \\
MIP & 17 & 23.6 & 4.0 & 2.6 & \textbf{2.3} & \textbf{0.62} & 0.79 & 0.94 & 0.70 \\
MG & 15 & 21.9 & 4.5 & 4.1 & \textbf{3.4} & 1.35 & 1.47 & 1.73 & \textbf{1.21} \\
MO & 16 & 19.0 & 5.7 & 5.7 & \textbf{4.6} & 1.15 & 0.98 & 1.30 & \textbf{0.89} \\
EMH & 5 & 9.1 & 8.1 & 8.1 & \textbf{7.0} & 0.89 & 1.03 & 0.91 & \textbf{0.88} \\
EV1 & 3 & 5.1 & 5.9 & 6.2 & \textbf{3.8} & \textbf{1.22} & 1.33 & 2.11 & 1.32 \\
EV2 & 3 & 7.8 & 6.9 & 7.9 & \textbf{5.8} & \textbf{1.23} & 1.63 & 2.91 & 1.48 \\
TC & 5 & 39.7 & 11.1 & 9.4 & \textbf{8.4} & 6.77 & 2.20 & 1.55 & \textbf{1.41} \\
TM & 6 & 183.1 & 93.4 & 91.2 & \textbf{90.7} & 18.95 & 10.93 & 10.55 & \textbf{10.52} \\
TO & 8 & 6071.0 & 5410.9 & 5397.0 & \textbf{5348.9} & \textbf{405.71} & 559.15 & 554.75 & 551.91 \\
TR & 6 & 8.7 & 3.7 & \textbf{0.9} & \textbf{0.9} & 0.67 & 0.80 & \textbf{0.45} & \textbf{0.45} \\
TS & 3 & 102.5 & 83.2 & 82.4 & \textbf{82.2} & 11.79 & 10.38 & \textbf{10.01} & \textbf{10.01} \\
\midrule
\textit{All}$^\ast$ & \textit{95} & \textit{34.8} & \textit{14.1} & \textit{13.3} & \textit{\textbf{12.4}} & \textit{2.84} & \textit{2.12} & \textit{2.28} & \textit{\textbf{1.90}} \\
\bottomrule
\end{tabular}
\begin{tablenotes}[para,flushleft]
\scriptsize
\textit{\textbf{BSL}~Basalt \quad \textbf{+LC}~loop closing \quad \textbf{+GBA}~global BA \quad \textbf{RS}~RootSLAM \\ \textbf{\#}~sequences per group \quad $^\ast$~excludes the outlying TO}
\end{tablenotes}
\end{threeparttable}%
}
\end{table}

We then verify the claim that motivates the square-root formulation, by running the same global bundle adjustment in double precision and in the single precision it enables. Table~\ref{tab:numerical-ablation} shows that accuracy is preserved, as the two agree within a centimeter of ATE in every group except TUM-VI outdoors, whose errors are two orders of magnitude larger, and neither precision is systematically better in ATE or RTE. The optimization, however, runs $2.16\times$ faster on average across the 103 sequences, ranging from $1.20\times$ on the slides group to $2.54\times$ on the outdoors one. Single precision therefore more than halves the cost of the global optimization at no measurable loss in accuracy.

\begin{table}[ht]
\centering
\resizebox{\columnwidth}{!}{%
\begin{threeparttable}
\caption{Numerical precision of global optimizer}
\label{tab:numerical-ablation}
\setlength{\tabcolsep}{3.5pt}
\begin{tabular}{@{}l r rr | rr | rr@{}}
\toprule
& & \multicolumn{2}{c|}{\textbf{ATE [cm]}} & \multicolumn{2}{c|}{\textbf{RTE [cm]}} & \multicolumn{2}{c@{}}{\textbf{Total time* [s]}} \\
\cmidrule(lr){3-4} \cmidrule(lr){5-6} \cmidrule(l){7-8}
\textbf{Sequence}
& \textbf{\#}
& \textbf{double} & \textbf{float}
& \textbf{double} & \textbf{float}
& \textbf{double} & \textbf{float} \\
\midrule
MIO & 16 & \textbf{7.0} & \textbf{7.0} & 1.27 & \textbf{1.26} & 4.59 & \textbf{2.03} \\
MIP & 17 & \textbf{2.3} & \textbf{2.3} & 0.71 & \textbf{0.70} & 15.19 & \textbf{7.74} \\
MG & 15 & 3.5 & \textbf{3.4} & \textbf{1.21} & \textbf{1.21} & 2.85 & \textbf{1.39} \\
MO & 16 & 4.9 & \textbf{4.6} & 0.91 & \textbf{0.89} & 3.68 & \textbf{2.07} \\
EMH & 5 & \textbf{7.0} & \textbf{7.0} & \textbf{0.88} & \textbf{0.88} & 2.69 & \textbf{1.29} \\
EV1 & 3 & \textbf{3.8} & \textbf{3.8} & \textbf{1.32} & \textbf{1.32} & 2.34 & \textbf{1.20} \\
EV2 & 3 & \textbf{5.7} & 5.8 & \textbf{1.40} & 1.48 & 2.70 & \textbf{1.38} \\
TC & 5 & \textbf{8.4} & \textbf{8.4} & 1.45 & \textbf{1.41} & 28.52 & \textbf{13.20} \\
TM & 6 & \textbf{90.0} & 90.7 & \textbf{10.45} & 10.52 & 69.56 & \textbf{33.62} \\
TO & 8 & 5352.9 & \textbf{5348.9} & 552.31 & \textbf{551.91} & 86.89 & \textbf{34.15} \\
TR & 6 & \textbf{0.9} & \textbf{0.9} & \textbf{0.45} & \textbf{0.45} & 1.47 & \textbf{0.80} \\
TS & 3 & \textbf{82.2} & \textbf{82.2} & \textbf{10.01} & \textbf{10.01} & 19.01 & \textbf{15.83} \\
\midrule
\textit{All} & \textit{103} & \textit{427.2} & \textit{\textbf{426.9}} & \textit{44.65} & \textit{\textbf{44.62}} & \textit{17.31} & \textit{\textbf{8.01}} \\
\bottomrule
\end{tabular}
\begin{tablenotes}[para,flushleft]
\scriptsize\raggedright
\textit{*Average GBA time on Intel Core i5-13600 \quad \textbf{\#}~sequences per group}
\end{tablenotes}
\end{threeparttable}%
}
\end{table}

Table~\ref{tab:sota-accuracy} compares RootSLAM against Basalt, OKVIS2 and ORB-SLAM3. The last two were run three times per sequence, keeping the one with the most successful frames, and breaking ties by the lowest ATE and then the lowest RTE. RootSLAM obtains the best ATE of the four systems in every MSD group. Both SLAM references degrade there by one or two orders of magnitude in the worst cases, such as MIO, where OKVIS2 reaches 451.7\,cm and ORB-SLAM3 398.9\,m against our 7.0\,cm.

On EuRoC and TUM-VI, the stereo datasets on which both baselines have been extensively tuned, they obtain the lowest ATE in most groups, but the three SLAM systems stay within a few centimeters of each other and RootSLAM remains below the decimeter in EMH, EV1, EV2, TC and TR. The TO sequences start in a room, run through open spaces for most of their length, and return to that room at the end, the only part with ground truth. The loops that reduce the most drift are the ones from this return to the initial room, where the views overlap little and not every system closes it. The comparison there is therefore closer to one between the VIO cores of the four systems, which accumulate tens of meters of drift.

\begin{figure}[!htbp]
   \centering
   \includegraphics[width=\linewidth]{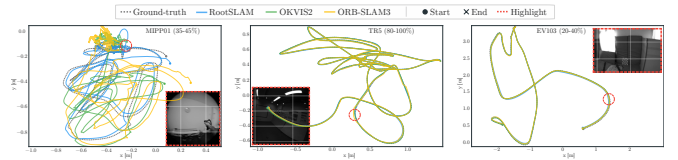}
   \caption{Estimated trajectories against ground truth on three segments of MIPP01, TR5 and EV103, with a frame of the scene from a highlighted section. RootSLAM stays clearly ahead on the MSD segment, where the fast head motion of MIPP01 degrades both references.}
   \label{fig:trajectories}
\end{figure}

Regarding success, Basalt, RootSLAM and OKVIS2 report an estimate for every camera frame of the 103 sequences, whereas ORB-SLAM3 averages 94.1\%. It is worst on MIO, where it goes down to 87.1\% of frames. This also makes its errors optimistic, since they are computed only over the frames it did track. Figure~\ref{fig:trajectories} illustrates this qualitatively on three segments, where RootSLAM follows the ground truth as closely as the references while being, as we show next, an order of magnitude faster than OKVIS2.

\begin{table}[ht]
\centering
\resizebox{\columnwidth}{!}{%
\begin{threeparttable}
\caption{Accuracy against the state of the art}
\label{tab:sota-accuracy}
\setlength{\tabcolsep}{2.5pt}
\begin{tabular}{@{}l r rrrr | rrrr | r@{}}
\toprule
& & \multicolumn{4}{c|}{\textbf{ATE [cm]}} & \multicolumn{4}{c|}{\textbf{RTE [cm]}} & \multicolumn{1}{c@{}}{\textbf{Succ.}} \\
\cmidrule(lr){3-6} \cmidrule(lr){7-10} \cmidrule(l){11-11}
\textbf{Seq.}
& \textbf{\#}
& \textbf{BSL} & \textbf{RS} & \textbf{OK2} & \textbf{OS3}
& \textbf{BSL} & \textbf{RS} & \textbf{OK2} & \textbf{OS3}
& \textbf{OS3}$^\dagger$ \\
\midrule
MIO & 16 & 33.0 & \textbf{7.0} & 451.7 & 39890 & 1.38 & \textbf{1.26} & 5.97 & 184.1 & 87.1 \\
MIP & 17 & 23.6 & \textbf{2.3} & 2.8 & 42.7 & 0.62 & 0.70 & \textbf{0.45} & 1.63 & 94.2 \\
MG & 15 & 21.9 & \textbf{3.4} & 10.5 & 827.1 & 1.35 & 1.21 & \textbf{0.89} & 11.54 & 98.4 \\
MO & 16 & 19.0 & \textbf{4.6} & 42.9 & 21.7 & 1.15 & 0.89 & \textbf{0.83} & 7.89 & 91.5 \\
EMH & 5 & 9.1 & 7.0 & \textbf{3.2} & 3.4 & 0.89 & \textbf{0.88} & 0.94 & 0.97 & 91.8 \\
EV1 & 3 & 5.1 & 3.8 & \textbf{2.2} & 2.6 & 1.22 & 1.32 & \textbf{1.14} & 1.26 & 95.0 \\
EV2 & 3 & 7.8 & 5.8 & \textbf{1.6} & 2.3 & 1.23 & 1.48 & \textbf{0.54} & 0.59 & 96.6 \\
TC & 5 & 39.7 & 8.4 & 5.4 & \textbf{2.6} & 6.77 & 1.41 & 1.00 & \textbf{0.75} & 97.4 \\
TM & 6 & 183.1 & 90.7 & 102.2 & \textbf{45.1} & 18.95 & 10.52 & 9.76 & \textbf{4.58} & 100.0 \\
TO & 8 & 6071 & \textbf{5349} & 370587 & 32623 & \textbf{405.7} & 551.9 & 20626 & 2247 & 100.0 \\
TR & 6 & 8.7 & 0.9 & \textbf{0.7} & 0.9 & 0.67 & \textbf{0.45} & 0.51 & 0.70 & 96.9 \\
TS & 3 & 102.5 & 82.2 & 57.0 & \textbf{33.7} & 11.79 & 10.01 & 6.85 & \textbf{4.03} & 100.0 \\
\midrule
\textit{All}$^\ast$ & \textit{95} & \textit{34.8} & \textit{\textbf{12.4}} & \textit{94.3} & \textit{6865} & \textit{2.84} & \textit{\textbf{1.90}} & \textit{2.39} & \textit{35.05} & \textit{94.1} \\
\bottomrule
\end{tabular}
\begin{tablenotes}[para,flushleft]
\scriptsize\raggedright
\textit{*~excludes TO \quad \textbf{BSL}~Basalt \quad \textbf{RS}~RootSLAM \quad \textbf{OK2}~OKVIS2 \quad \textbf{OS3}~ORB-SLAM3 \\ \textbf{Success}~frames with an estimate \quad \textbf{\#}~sequences per group \quad $^\dagger$ 100\% for the rest}
\end{tablenotes}
\end{threeparttable}%
}
\end{table}

\begin{table}[t]
\centering
\resizebox{\columnwidth}{!}{%
\begin{threeparttable}
\caption{System times on representative sequences}
\label{tab:timing}
\setlength{\tabcolsep}{6.6pt}
\begin{tabular}{@{}l r rrrr | rr@{}}
\toprule
& & \multicolumn{4}{c|}{\textbf{Mean frame time [ms]}} & \multicolumn{2}{c@{}}{\textbf{Global BA [s]}} \\
\cmidrule(lr){3-6} \cmidrule(l){7-8}
\textbf{Sequence}
& \textbf{Dur. [s]}
& \textbf{BSL} & \textbf{RS} & \textbf{OK2} & \textbf{OS3}
& \textbf{RS} & \textbf{OK2} \\
\midrule
EMH02 & 152.0 & 11.8 & \textbf{11.5} & 191.5$^\dagger$ & 40.9 & \textbf{2.83} & 6.35 \\
EV102 & 85.5 & \textbf{8.9} & 9.3 & 132.5$^\dagger$ & 38.9 & \textbf{1.22} & 6.07 \\
EV202 & 117.4 & 9.4 & \textbf{9.0} & 132.6$^\dagger$ & 40.3 & \textbf{3.26} & 6.87 \\
\textit{Group Mean} & \textit{118.3} & \textit{10.1} & \textit{\textbf{9.9}} & \textit{152.2}$^\dagger$ & \textit{40.0} & \textit{\textbf{2.44}} & \textit{6.43} \\
\midrule
MIO02 & 87.3 & 15.2 & \textbf{15.1} & 80.7$^\dagger$ & 40.0$^\dagger$ & \textbf{0.77} & 13.16 \\
MIPB02 & 221.3 & \textbf{15.6} & 15.8 & 116.2$^\dagger$ & 51.4$^\dagger$ & \textbf{6.03} & 12.85 \\
MGO02 & 157.2 & \textbf{11.9} & 12.1 & 184.8$^\dagger$ & 32.2 & \textbf{1.43} & 47.39 \\
MOO02 & 179.9 & \textbf{7.0} & 7.3 & 116.9$^\dagger$ & 34.9$^\dagger$ & \textbf{0.90} & 73.32 \\
\textit{Group Mean} & \textit{161.4} & \textit{\textbf{12.4}} & \textit{12.6} & \textit{124.6}$^\dagger$ & \textit{39.6}$^\dagger$ & \textit{\textbf{2.28}} & \textit{36.68} \\
\midrule
TC1 & 299.5 & \textbf{8.8} & 9.8 & 108.1$^\dagger$ & 38.3 & \textbf{18.43} & 22.28 \\
TM1 & 772.3 & \textbf{9.6} & 9.8 & 199.9$^\dagger$ & 44.5 & \textbf{5.03} & 86.19 \\
TO2 & 922.4 & \textbf{10.2} & 10.8 & 205.4$^\dagger$ & 37.5 & \textbf{25.74} & 94.01 \\
TR6 & 131.8 & \textbf{9.3} & 9.8 & 123.0$^\dagger$ & 34.4 & \textbf{1.22} & 2.05 \\
TS1 & 279.1 & 10.0 & \textbf{9.6} & 132.9$^\dagger$ & 33.8 & \textbf{7.07} & 24.30 \\
\textit{Group Mean} & \textit{481.0} & \textit{\textbf{9.6}} & \textit{10.0} & \textit{153.9}$^\dagger$ & \textit{37.7} & \textit{\textbf{11.50}} & \textit{45.77} \\
\midrule
\textit{Total Mean} & \textit{283.8} & \textit{\textbf{10.6}} & \textit{10.8} & \textit{143.7} & \textit{38.9} & \textit{\textbf{6.16}} & \textit{32.90} \\
\bottomrule
\end{tabular}
\begin{tablenotes}[para,flushleft]
\scriptsize\raggedright
\textit{\textbf{BSL}~Basalt\quad\textbf{RS}~RootSLAM\quad\textbf{OK2}~OKVIS2\quad\textbf{OS3}~ORB-SLAM3\quad\textbf{Dur.}~sequence duration \quad \textbf{Global~BA}~optimization time on Intel Core Ultra 9 285H \quad $^\dagger$above real-time budget}
\end{tablenotes}
\end{threeparttable}%
}
\end{table}

\begin{figure}[!htbp]
    \centering
    \includegraphics[width=\linewidth]{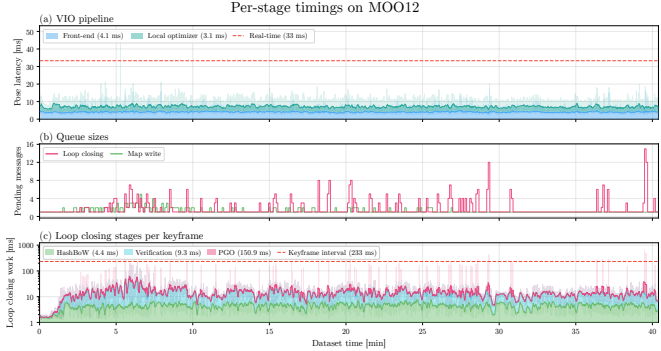}
    \caption{Per-stage timings of RootSLAM on MOO12, a 42\,min stereo sequence at
   30\,Hz, run on an Intel Core Ultra 9 285H. (a)~Latency of the VIO pipeline, accounting for front-end and local optimizer threads, needed to estimate a pose. (b)~Pending messages of the queues feeding the loop closer and the map. (c)~Work spent per keyframe on each loop closing stage, in log scale. Solid lines are a moving average over a 6\,s window, with the raw data in the background. For (b) we show the peak size of each window. In parentheses, the mean time of each stage.}
    \label{fig:timings-per-stage}
\end{figure}

We finally evaluate the speed of the system. Figure~\ref{fig:timings-per-stage} breaks the processing time down per stage on the 42-minute sequence MOO12, with the loop closing and map components running asynchronously in the background. The VIO pipeline, which determines when a pose is delivered, averages 7.2\,ms, comfortably below the 33\,ms budget of its frame rate. The local optimizer takes a keyframe every 7 frames, leaving an interval of 233\,ms to process each one. The queues feeding the loop closer and the map peak at 15 pending keyframes, but drain almost immediately. Pose-graph optimization is the most expensive stage at 150.9\,ms on average, but it runs only on the keyframes that close a loop above the drift threshold, and still fits within the keyframe interval.

\begin{figure}[!htbp]
    \centering
    \includegraphics[width=\linewidth]{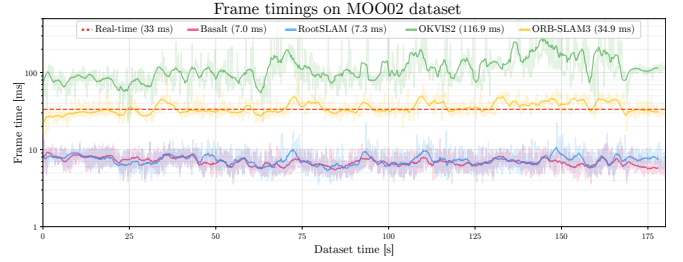}
    \caption{Per-frame processing time on MOO02, from MSD at \SI{30}{\hertz}, in log scale. Solid lines are a moving average over a 2\,s window, with the raw data in the background. The dashed line is the real-time budget of the sequence. RootSLAM tracks the Basalt VIO it extends within a fraction of a millisecond and stays far below the budget.}
    \label{fig:timings-sota}
\end{figure}

Figure~\ref{fig:timings-sota} compares the per-frame processing time against the baselines on MOO02. RootSLAM stays essentially at the same frame times as Basalt and far below the real-time budget, whereas ORB-SLAM3 hovers around it, crossing it systematically, and OKVIS2 exceeds it by more than 3$\times$ on average. Table~\ref{tab:timing} quantifies this over twelve sequences, one per dataset category, since sequences in the same group show similar run times. RootSLAM processes a frame in 10.8\,ms on average, $13\times$ faster than OKVIS2 and $3.6\times$ faster than ORB-SLAM3, and is, together with the VIO it extends, the only system whose mean frame time stays below the real-time budget of every sequence. Its global optimization is also $5.3\times$ faster than the full bundle adjustment of OKVIS2 on average. The gap widens with the length of the sequence, because on TM1, which is 772.3\,s long, RootSLAM processes every frame plus its global optimization in 157.2\,s, $4.9\times$ faster than real time, while OKVIS2 needs 3173.0\,s, $20\times$ slower than RootSLAM and $4.1\times$ slower than real-time. This follows from the per-frame cost growth that OKVIS2 exhibits over long sessions, which RootSLAM avoids by bounding the size of its map.
\section{CONCLUSIONS}

We introduced RootSLAM, a fully-featured multi-camera visual-inertial SLAM system that extends the VIO from Basalt with the necessary components for long-term operation. It adds a persistent map with keyframe culling, loop detection and pose-graph optimization, and a global bundle adjustment stage. We propose a reformulation of the SfM-focused square-root bundle adjustment method suited for the global optimization of a SLAM system. This formulation enables the global optimizer to run entirely in single precision floating point arithmetic, roughly halving its runtime with no loss in accuracy. Furthermore, a rig-based parametrization ensures the formulation's validity for arbitrary multi-camera configurations and camera models.

An ablation study isolates the contribution of each component and verifies the speedup of the square-root global optimizer over its double-precision counterpart. Our evaluation on the Monado SLAM Dataset, EuRoC, and TUM-VI demonstrates that RootSLAM achieves comparable absolute and relative trajectory errors to the state-of-the-art works ORB-SLAM3 and OKVIS2, while running up to $20\times$ faster, staying below real-time at all times, and avoiding the per-frame cost growth exhibited by OKVIS2 over long sessions. The performance gains of RootSLAM make it feasible to deploy the system in ever cheaper, cooler, and power-efficient devices in need of spatial understanding.

Upon acceptance, we will release the full implementation, encompassing map management, loop detection, pose-graph optimization, and global mapping modules, to the open-source community.

\bibliographystyle{IEEEtran}
\balance
\bibliography{IEEEabrv,main}

\immediate\closein\imgstream

\end{document}